# The Application of Preconditioned Alternating Direction Method of Multipliers in Depth from Focal Stack

Hossein Javidnia[1], *Student, IEEE*, Peter Corcoran[2], *Fellow, IEEE*

## Abstract

Post capture refocusing effect in smartphone cameras is achievable by using focal stacks. However, the accuracy of this effect is totally dependent on the combination of the depth layers in the stack. The accuracy of the extended depth of field effect in this application can be improved significantly by computing an accurate depth map which has been an open issue for decades. To tackle this issue, in this paper, a framework is proposed based on Preconditioned Alternating Direction Method of Multipliers (PADMM) for depth from the focal stack and synthetic defocus application. In addition to its ability to provide high structural accuracy and occlusion handling, the optimization function of the proposed method can, in fact, converge faster and better than state of the art methods. The evaluation has been done on 21 sets of focal stacks and the optimization function has been compared against 5 other methods. Preliminary results indicate that the proposed method has a better performance in terms of structural accuracy and optimization in comparison to the current state of the art methods.

## 1 Introduction

The compact design of mobile cameras does not allow users access to lens properties such as the aperture. By having the control over the aperture in a camera, one can control the camera's depth of field (and the light flux entering the camera). This means the user can decide how much of an image remains in focus around an object.

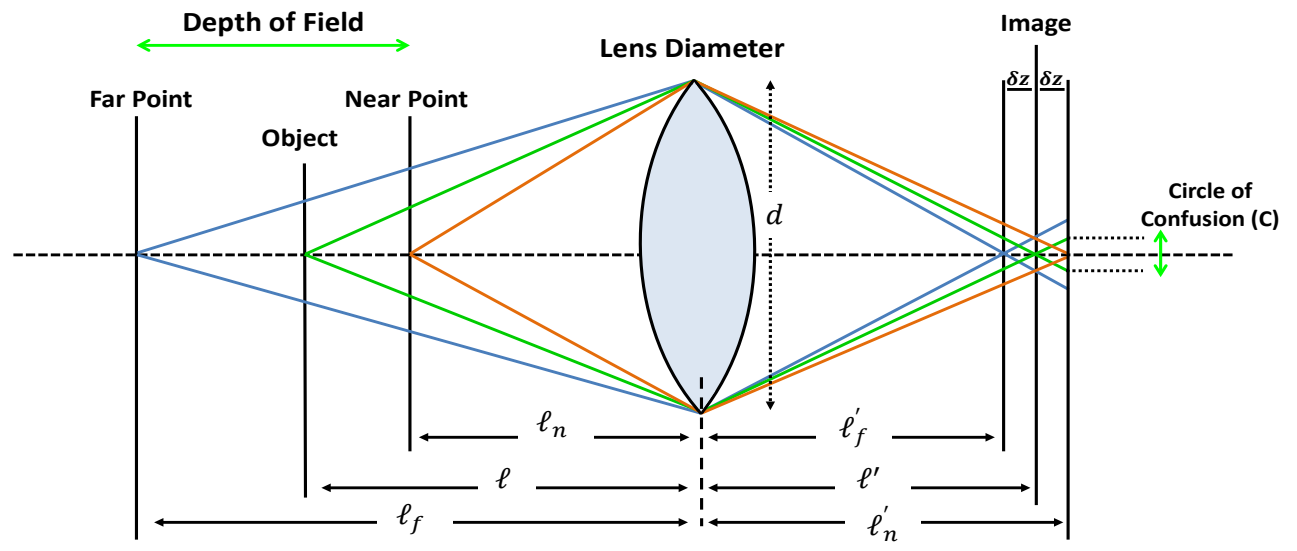

a. Depth of field (DoF) and circle of confusion for a lens representing light collection optics in an imaging system

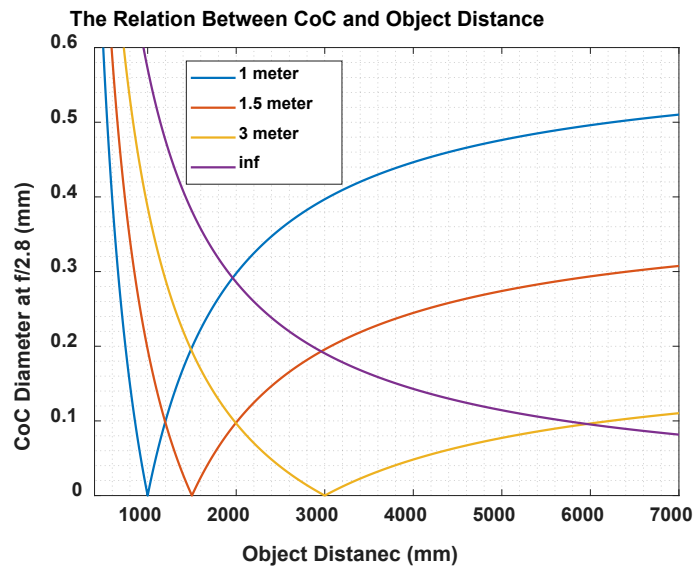

b. The relation between Circle of Confusion and the object distance/focus position

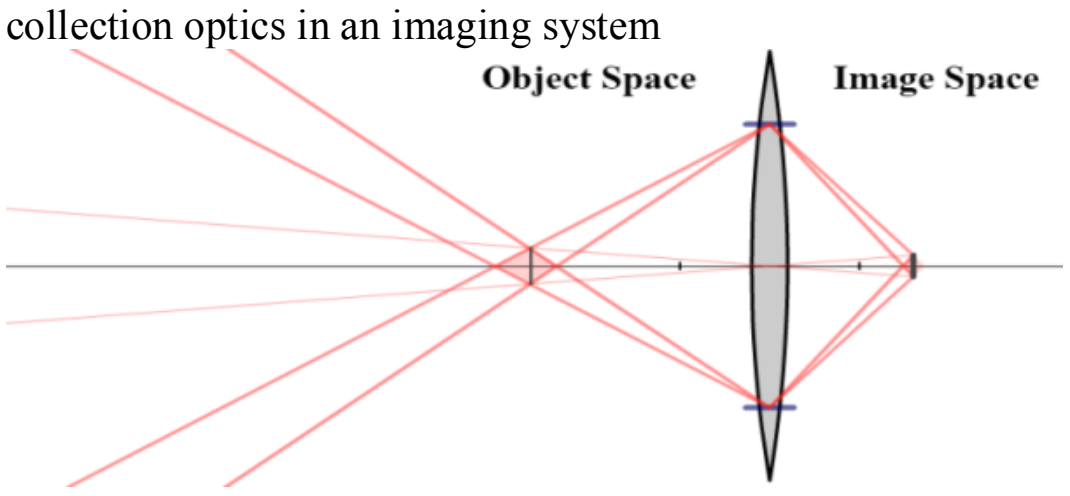

c. Small depth of field with F-number 0.4

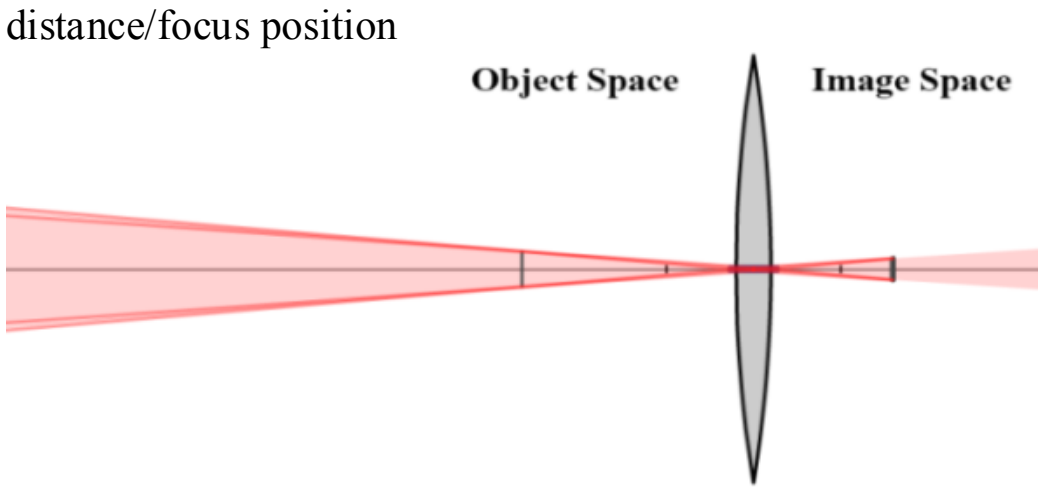

d. Large depth of field with F-number 32

**Figure 1. Demonstration of the relation between F-number and depth of field, Circle of Confusion and Object Distance**

---

The research work presented here was funded under the Strategic Partnership Program of Science Foundation Ireland (SFI) and co-funded by SFI and FotoNation Ltd. Project ID: 13/SPP/I2868 on "*Next Generation Imaging for Smartphone and Embedded Platforms*".

[1] H. Javidnia is with the Department of Electronic Engineering, College of Engineering, National University of Ireland, Galway, University Road, Galway, Ireland. (e-mail: h.javidnia1@nuigalway.ie).

[2] P. Corcoran is with the Department of Electronic Engineering, College of Engineering, National University of Ireland, Galway, University Road, Galway, Ireland. (e-mail: peter.corcoran@nuigalway.ie).

Fig.1 (a) shows schematically the relation between the depth of focus (in image space) and depth of field (in object space). As illustrated in Fig.1 (c), small depth of field will make the main object in focus, while the rest of the image will be less sharp. A large depth of field will keep the entire image sharp throughout its depth; this concept is shown in Fig.1 (d). When light rays from an out of focus point source enter a lens, the point on the object is focused into a circle on the image plane. This circle is called the Circle of Confusion (CoC) which is shown as C in Fig.1 (a). The size of the CoC in used to measure the sharpness of an image. The bigger CoC shows that the point on the object is more out of focus. The diameter of CoC depends on focal length $\boldsymbol{f}$, object distance $\boldsymbol{\ell_n}$(near point) and the distance between the object point and the lens $\boldsymbol{\ell}$ and aperture diameter $\boldsymbol{d}$. Therefore, the diameter of the CoC can be calculated using Eq. 1:

$$\boldsymbol{C} = \frac{\boldsymbol{d f |\ell - \ell_n|}}{\boldsymbol{\ell(\ell_n - f)}} \qquad (\mathbf{1})$$

Fig.1 (b) illustrates the relationship between the CoC and the object distance for an aperture f/2.8 mm for a specific camera model. The *X*-axis in Fig.1 (b) represents the distance of the object points in focus and the *Y*-axis shows how far the object would be in focus. For instance, if an object is located at the distance 1000 mm the value of CoC is 0, this means the object is fully in focus. If an object is located at the distance beyond 1200 mm then the 1.5 meter setting of the camera should be used. If an object is located at the distance beyond 2000 mm then the 3 meter setting of the camera should be used and for the objects which are located at the distance beyond 6000 mm, the infinite setting of the camera should be used.

The adjustable aperture feature is available in DSLR cameras but smartphone cameras have a fixed aperture as they are designed for ease of portability, robustness and low cost.

To overcome this shortcoming, post-capture image refocusing can be employed by using Depth from Focus (DfD) [1, 2] and focal stack. The focal stack is a collection of images with different focus points which correspond to different depth layers. The focal setting presenting the maximum sharpness of pixel $\boldsymbol{p}$ corresponds to the depth of the pixel or its distance to the camera. The combination of these images can generate the extended depth of field similar to the range being generated by optical properties of the camera. The accuracy of this effect is highly dependent on the accuracy of the corresponding depth map.

In handheld devices such as smartphones, a focal stack is generated by automatic focal plane sweeps to find the camera's best auto-focus setting while taking photos. However, in dynamic scenes, the slight translation of the camera by users and their handshake can introduce motion parallax. The experiments indicate that it takes about $\frac{1}{2}$ to $\frac{1}{3}$ seconds for a camera to capture the full extent of its focus setting. This means that the local parallax met within this short time frame can be dismissed but yet an alignment procedure is required to compensate the parallax between the first and last frame of the focal stack. Generally, an aligned focal stack should be similar to a stack captured by a telecentric camera.

In this paper, we present a framework to compute and optimize the dense depth map from the high-resolution focal stack which can be used to produce an accurate synthetic defocus. The framework initiates by taking a focal stack from a moving camera as the input and generating a stabilized image sequence. At the second step, the initial depth map is estimated from the stabilized focal stack. At the end, Preconditioned Alternating Direction Method of Multipliers (PADMM) with a new cost function is applied to refine depth discontinuities and generate a noise-free depth map.

The proposed method has several advantages in comparison to the state of the art methods, such as:

1- Fast and better convergence of the optimization function
2- Occlusion handling in the generated depth map
3- High structural accuracy of the depth map
4- High performance in texture-less scenes
5- Accurate depth information along objects' boundaries and surface

The rest of this paper is organized as follows: Section II outlines the previous researches. The proposed method is explained in detail in Section III. The evaluation results are presented in Section IV and Section V includes conclusion and feature work.

## 2 Previous Work

A considerable amount of researches focused on depth from focus control for decades [3-8]. Most of these methods concentrated on depth from focus/defocus or depth recovery from the focal stack on light field cameras [4, 5, 9, 10]. Using light field cameras has an advantage of capturing simultaneous multiple views with variable focal points which provide more accurate information about the depth of the scene; however, the images are captured in low resolution and in small aperture the value of Signal-to-noise ratio (SNR) is significantly low [4]. The size of these cameras along with the mentioned challenges makes them inapplicable for handheld devices such as smartphones. Another disadvantage of light field cameras is the disability in handling occlusion due to the lack of lateral variation being captured in different viewpoints [9].

A framework to recover depth from the focal stack is presented in [3] to handle images captured on smartphones. The focal stack is being aligned to make it as similar as a focal stack captured by a telecentric camera. Multilabel Markov Random Field (MRF) optimization is used to generate all in focus image from the aligned stack. This method works quite well for the Lambertian scenes however the optimization problem during the calibration process is highly non-convex and that makes this process considerably slow. The other problem with this method is the processing time of the non-linear least squares minimization to jointly optimize the initially estimated aperture size, focal depths, focal length and the depth map. The whole processing time of this method is ~20 minutes for 25 frames with 640×360 pixels resolution which make this algorithm almost inapplicable as a smartphone application. The depth maps generated by this method suffer from inaccurate depth values on objects surface, especially on reflective surfaces. In some cases, the depth information along the boundaries of the foreground object is mixed with the values on the background and that might result in an inaccurate synthetic defocus.

The complexity of the non-convex optimization is reformulated in [6] where depth from focus is presented as a variational problem by introducing a nonconvex data fidelity term and a convex nonsmooth regularization. The nonconvex minimization problem in [6] is aimed to be solved by a linearized alternating directions method of multipliers. This method has a superior performance in comparison to state of the art methods but the convergence of the optimization function happens very slowly and in a high number of iterations. Also, the depth map generated by this method suffers from inaccurate depth values on objects surface and missing edges and corners.

Some other approaches in this field have been proposed to facilitate the depth from focus applications by introducing coded focal stack photography [11] or coded aperture photography [12, 13]. These methods require physical changes in the structure of the camera and yet the generated depth maps suffer from lack of structural quality.

Persch *et al.* [14] proposed a variational approach for the problem of depth from defocus based on modeling of the image formation by featuring the thin lens model and preserving the crucial physical properties such as maximum-minimum principle for the intensity values. Later, the variational model is minimized using the multiplicative Euler–Lagrange. The proposed solution in [14] appears to generate false depth levels in relatively close scenes and in general, the depth profiles are likely to be affected by the color information as the robustification method employed in [14] uses the full-color information of the focal stack.

Pérez *et al.* [15] proposed a focal stack frequency decomposition algorithm from light field images based on the trigonometric interpolation principle as the discrete focal stack transform. The proposed method in [15] utilizes fast discrete Fourier transform to generate refocus planes in a reasonably fast computational time. The reverse of this transformation in studied in [16] where a focal stack is used to obtain a 4D light field image set using discrete focal stack transform.

Unlike [15], Mousnier *et al.* [17] presented an approach to reconstruct 4D light field image sets from a stack of images taken by a fixed camera at different focal points. The algorithm initiates by calculating the focus map by utilizing region expansion with graph cut. Later, the depth map is estimated based on the calibration details of the camera and it is used to reconstruct the Epipolar images. The reconstructed Epipolar images are used for refocusing purposes.

Bailey *et al.* [18] proposed a method to calculate depth from the focal stack by estimating the level of the blurriness for each pixel. The method initiates by applying a focus measure to each pixel in the stack. A normalized convolution is proposed to extrapolate the invalid blur estimates. Afterwards, the per-pixel depth is calculated based on the blur estimations.

Jeong *et al.* [19] presented a post-processing approach to refine the estimated depth map from two images captured with different focal points. The initial depth map is calculated using a depth from defocus algorithm. To improve the quality of the depth map, mean-shift clustering is applied to the first input image to obtain the segmented image. A single depth value is assigned to each segment of the image by averaging all depth values in the corresponding segment.

Surh *et al.* [20] presented a novel focus measure to determine how in focus a point is on an image. The shape of the focus measure introduced in [20] contains a disk which focuses on the pixel of interest and the ring which surrounds the disk. To estimate the depth map, the initial calculated cost volume is aggregated by employing tree-based cost aggregation method. Afterwards, the depth discontinuity and unreliable depth labels are filtered based on the median of absolute deviation map and using tree-based cost aggregation method.

Focal stacks are also used to handle some of the optical features such as post-capture perspective shift and aperture reshaping. Alonso [21] developed a method in the Fourier domain for post-capture aperture reshaping in focal stacks. This allows users to change the blur shape for the out of focus points. This study came to the conclusion that by utilizing domain transformation methods such as Fourier it is possible to manipulate the optical setting of the camera. In another study, Alonso *et al.* [22] proposed a method for post-capture perspective shift reconstruction of a 3D scene from a focal stack. Unlike the computational approaches which estimate the depth map, the method in [22] takes advantage of depth-variant point-spread function to introduce the lateral ($(x, y)$ plane) and axial ($z$ plane) shifts.

## 3 Proposed Method

### 3.1 Focal Stack Alignment

To generate the parallax-free input focal stack we refer to Epipolar homography alignment. To do that, we merge all the homographies into Epipolar geometry. Considering there are $\boldsymbol{j}$ plane patches in an image and their corresponding maps in the second image are characterized as:

$$\begin{aligned} H_1 &= s_1\mathcal{R}(I - \mathcal{T}N_1^{\mathcal{T}}) \\ H_2 &= s_2\mathcal{R}(I - \mathcal{T}N_2^{\mathcal{T}}) \\ &\dots \\ H_j &= s_j\mathcal{R}(I - \mathcal{T}N_j^{\mathcal{T}}) \end{aligned} \quad (2)$$

where $\boldsymbol{s}$ is a scale factor, $\mathcal{R}$ is a $3 \times 3$ rotation matrix, $\boldsymbol{I}$ is the image, $\mathcal{T}$ is the second camera's translation from first camera's point of view and $\boldsymbol{N}(\mathfrak{n}_1, \mathfrak{n}_2, \mathfrak{n}_3)$ is the normal vector of the plane surface. So we can write:

$$\frac{s_1}{s_i}H_i - H_1 = s_1\mathcal{R}\mathcal{T}N_1^{\mathcal{T}} - s_1\mathcal{R}\mathcal{T}N_i^{\mathcal{T}} = \mathcal{K}\Delta N_i^{\mathcal{T}} \quad (3)$$

$$\mathcal{K} = (\kappa_1\ \kappa_2\ \kappa_3)^{\mathcal{T}} = \mathcal{R}\mathcal{T}$$

where $\Delta N_i = (\Delta\mathfrak{n}_1\ \Delta\mathfrak{n}_2\ \Delta\mathfrak{n}_3)^{\mathcal{T}} = s_1(N_1 - N_i)$. Consequently it can be concluded that:

$$d_iH_i = H_1 + \mathcal{K}\Delta N_i^{\mathcal{T}} \qquad i = 2, 3, \dots, j \qquad (4)$$

where $\boldsymbol{d} = \frac{1}{||N||}$ is the distance of the plane from the origin and $\boldsymbol{H_1}$ is represent the correlation between the basis homography and all the other homographies. The important feature of the Eq. 4 is that it reduces the number of independent parameters of a homography and makes them equal to the degree of freedom of a system with $\boldsymbol{j}$ planar surface. Generally, a homography includes 5 *dof* indicating the camera motion and 3 *dof* representing the plane surface normal. Assuming more than one plane between two images, then $\boldsymbol{j}$ homographies will have $\boldsymbol{8j}$ parameters. Eq. 4 decreases the number of the parameters to $\boldsymbol{5 + 3j}$ which is equivalent of the total degree of freedom in a system with $\boldsymbol{j}$ planar surface.

Using Eq. 4 the motion estimation can break down into two parts:

First, considering that $\boldsymbol{H_1}$ and $\mathcal{K}$ are fixed, it is possible to characterize $\Delta N_i$ and $\boldsymbol{H_i}$ by utilizing least square algorithm for each plane patches. To estimate $\Delta N_i$ we define two vectors as:

$$\begin{aligned} \mathcal{V}_1 &= (\kappa_1x - \kappa_3xx' \quad \kappa_1y - \kappa_3yx' \quad \kappa_1 - \kappa_3x') \\ \mathcal{V}_2 &= (\kappa_2x - \kappa_3xy' \quad \kappa_2y - \kappa_3yy' \quad \kappa_2 - \kappa_3y') \end{aligned} \quad (5)$$

So $\Delta N_i$ can be estimated using least squares method as:

$$\begin{aligned} \mathcal{V}_1\Delta N_i &= x'(h_7x + h_8y + 1) - (h_1x + h_2y + h_3) = b_{1i} \\ \mathcal{V}_2\Delta N_i &= y'(h_7x + h_8y + 1) - (h_4x + h_5y + h_6) = b_{2i} \end{aligned} \quad (6)$$

where $\boldsymbol{h_{1-8}}$ are the parameters of the homography matrix. $(x, y, z)$ and $(x', y', z')$ are the coordinates of the point $\boldsymbol{P}$ in two camera frames.

The second part is somehow the inverse process of the first part. Assuming $\boldsymbol{\Delta N_i}$ is fixed, $\boldsymbol{H_1}$ and $\boldsymbol{\mathcal{K}}$ can be updated by utilizing another least square process. To estimate $\boldsymbol{H_1}$ and $\boldsymbol{\mathcal{K}}$ we define three vectors as:

$$
\begin{aligned}
\boldsymbol{E_i} &= (\boldsymbol{x}\ \boldsymbol{y}\ \mathbf{1}\ \mathbf{0}\ \mathbf{0}\ \mathbf{0}\ -\boldsymbol{xx'}\ \ -\boldsymbol{yx'}\ \ \boldsymbol{\Delta N_i P}\ \ \mathbf{0}\ \ -\boldsymbol{x'\Delta N_i P}) \\
\boldsymbol{F_i} &= (\mathbf{0}\ \mathbf{0}\ \mathbf{0}\ \boldsymbol{x}\ \boldsymbol{y}\ \mathbf{1}\ -\boldsymbol{xy'}\ \ -\boldsymbol{yy'}\ \ \mathbf{0}\ \ \boldsymbol{\Delta N_i P}\ \ -\boldsymbol{y'\Delta N_i P}) \qquad (7) \\
\boldsymbol{G} &= (\boldsymbol{h_1}\ \boldsymbol{h_2}\ \boldsymbol{h_3}\ \boldsymbol{h_4}\ \boldsymbol{h_5}\ \boldsymbol{h_6}\ \boldsymbol{h_7}\ \boldsymbol{h_8}\ \boldsymbol{k_1}\ \boldsymbol{k_2}\ \boldsymbol{k_3})
\end{aligned}
$$

where $\boldsymbol{P} = (\boldsymbol{x}, \boldsymbol{y}, \boldsymbol{z})$ is a point on the plane surface and $\boldsymbol{\Delta N_i P} = (\boldsymbol{\Delta n_{i1} x} + \boldsymbol{\Delta n_{i2} y} + \boldsymbol{\Delta n_{i3} z})$. So it can be concluded that $\boldsymbol{E_i G^{\mathcal{T}}} = \boldsymbol{x'}$ and $\boldsymbol{F_i G^{\mathcal{T}}} = \boldsymbol{y'}$. Then $\boldsymbol{G}$ can be estimated using least square process as:

$$\boldsymbol{G^{\mathcal{T}}} = \frac{\boldsymbol{\mathcal{B}}}{\boldsymbol{Q}} \qquad (8)$$

where $\boldsymbol{\mathcal{B}} = \begin{pmatrix} x'_{11} \\ y'_{11} \\ \vdots \\ x'_{jn} \\ y'_{jn} \end{pmatrix}$ and $\boldsymbol{Q} = \begin{pmatrix} E_{11} \\ F_{11} \\ \vdots \\ E_{jn} \\ F_{jn} \end{pmatrix}$. By estimating $\boldsymbol{\Delta N_i}$ from Eq. 6 and $\boldsymbol{H_1}$ and $\boldsymbol{\mathcal{K}}$ using Eq. 8, one can construct the global homography from Eq. 4. The alignment process will be over when the average reprojection error is smaller than a threshold.

### 3.2 Depth Estimation and Regularization

The depth estimation process starts with calculating the value of the focus factor for each pixel at every frame of the aligned focal stack. The value of the focus factor for a pixel $(\boldsymbol{i}, \boldsymbol{j})$ over all the frames in the stack is referred as focus function. The Modified Laplacian is used in this case to compute the focus function of $\boldsymbol{I}$:

$$\boldsymbol{\mathcal{F}_{\mathcal{Y}}} = (|\boldsymbol{I} \times \boldsymbol{C_x}| + |\boldsymbol{I} \times \boldsymbol{C_{\mathcal{Y}}}|) \times \boldsymbol{m_r} \qquad (9)$$

where the convolution masks on $\boldsymbol{x}$ and $\boldsymbol{\mathcal{Y}}$ domains are $\boldsymbol{C_x} = [-\mathbf{1}, \mathbf{2}, -\mathbf{1}]$ and $\boldsymbol{C_{\mathcal{Y}}} = \boldsymbol{C_x^T}$, respectively. The mean filter mask is used as $\boldsymbol{m}$ by the radius $\boldsymbol{r}$. The initial depth map is computed by modeling the focus function using the 3-point Gaussian distribution [23]. The algorithm relies on 3 focus factors $\boldsymbol{\mathcal{F}_{\mathcal{Y}-1}}$, $\boldsymbol{\mathcal{F}_{\mathcal{Y}}}$ and $\boldsymbol{\mathcal{F}_{\mathcal{Y}+1}}$. This will result the following focus function:

$$\boldsymbol{\mathcal{F}} = \boldsymbol{\mathcal{F}_{max}} \boldsymbol{exp}\left\{-\frac{(\boldsymbol{M} - \boldsymbol{S})^2}{2\sigma_{\mathcal{F}}^2}\right\} \qquad (10)$$

where $\boldsymbol{S}$ and $\boldsymbol{\sigma_{\mathcal{F}}}$ are the mean standard deviation of the Gaussian distribution and $\boldsymbol{M}$ is the displacement of the object plane. The estimated depth values correspond to the location of $\boldsymbol{\mathcal{F}_{max}}$. As long as there is a good correlation between the Gaussian model and the focus function, the depth values get more authentic. But this situation is not constant and it can be interrupted by a variety of reasons such as noise. The presence of noise in the image domain can cause the focus function not to fit on the Gaussian model. That means the initial depth map is suffering from uncertain depth values. This condition becomes severe in case of small motions of the camera. Fig.2 (b) shows the initial estimated depth map.

This problem is reformulated to a convex minimization problem to be solved by PADMM [24, 25]. To define the formulation of the convex problem we refer to regularization method proposed by Rudin, Osher and Fatemi (ROF) [26] which introduces a minimization problem to generate the restored image $\boldsymbol{t}$ for a noisy image $\boldsymbol{I}$ as:

$$P(x) = \frac{\mathcal{N}^2}{2} \times \lambda + \mathrm{K} \qquad (11)$$

where $\lambda$ is the regularization parameter and:

$$\mathcal{N} = \int_{\mathrm{F}} (I(x) - t(x)) \ dx, \qquad (12)$$

$$I: \mathrm{F} \rightarrow \mathbb{R} \text{ (F is bounded open subset of } \mathbb{R}^2\text{)}$$

and **K** defines the vectorial gradient as:

$$\mathrm{K} = \mathbf{sup}\{\mathrm{B}: \vec{\mathcal{G}} \in (\mathrm{F}, \mathbb{R}^2)^2\} \qquad (13)$$

$$\mathrm{B} = \int_{\mathrm{F}} t(x) div\vec{\mathcal{G}}(x) dx \qquad (14)$$

**K** prevents the function to generate ringing artifacts along the edges but it cannot handle the discontinuities. It generally presents a loss of contrast which happens due to the use of $\ell^2$ fidelity. To overcome this issue, the ROF function is changed to a unique global minimizer by employing the vectorial $\ell^1$ norm fidelity term.

$$P(x) = \mathcal{N} \times 2^{\lambda} + \mathrm{K} \qquad (15)$$

Using the $\ell^1$ norm allows to solve non-convex optimization problems using convex optimization methods. The important advantage of using the convex optimization is that the global optimum is achievable with a high precision in a shorter computational time. It is also independent from the initialization.

Since the problem can be solved using convex optimization, we attempt to solve the ROF minimization by modifying PADMM constrained convex minimization method. Consider a generic constraint minimization problem as:

$$(p, q) = \mathbf{arg}\, min_{(p,q)}\{R(p) + S(q) \text{ subject to } T(p, q) = l\} \qquad (16)$$

where $R$ and $S$ are proper, closed convex functions, $T$ denote a nonlinear operator and $l$ is the specified function.

Eq. 16 is solved by alternating minimization of the augmented Lagrange function:

$$\mathcal{L}_{\daleth} = R(p) + S(q) + \langle \rho, T(p, q) - l \rangle + \frac{\daleth \|T(p, q) - l\|_2^2}{2} \qquad (17)$$

where $(p, q)$ are the solution vectors, $\rho$ is a sequence of estimates of the Lagrange multipliers of the constraints $T(p, q) = l$ and $\daleth > 0$ is a predefined penalty parameter. Giving the residuals as $r = T(p, q) - l$ and the dual variable as $\rho$, we can express the ADMM problem as:

$$p^{k+1} \in arg\,min_p \left\{ R(p) + \langle \rho^k, T(p, q^k) \rangle + \frac{\daleth \left\| T(p, q^k) - l \right\|_2^2}{2} \right\} \quad (18)$$

$$q^{k+1} \in arg\,min_q \left\{ S(q) + \langle \rho^k, T(p^{k+1}, q) \rangle + \frac{\daleth \left\| T(p^{k+1}, q) - l \right\|_2^2}{2} \right\} \quad (19)$$

$$\rho^{k+1} = \daleth (T(p^{k+1}, q^{k+1}) - l) + \rho^k \quad (20)$$

where $k$ is the iteration number. By finding the linear approximation of $T(p^{k+1}, q^k)$ and $T(p^{k+1}, q^{k+1})$ around $p^k$ and $q^k$ using the Taylor expansion, we can reduce the nonlinearity computation overhead of Eq. 18 and Eq. 19. So:

$$T(p, q^k) \cong T(p^k, q^k)\left(1 + \vartheta_p (p - p^k)\right) \quad (21)$$

$$T(p^{k+1}, q) \cong T(p^{k+1}, q^k)\left(1 + \vartheta_q (q - q^k)\right) \quad (22)$$

To convert ADMM to a preconditioned solver, we modify Eq. 18 and Eq. 19 by adding an additional proximity term as:

$$\frac{\lambda \left\| p^{k+1} - p^k \right\|_{Z_1^k}^{2^{\lambda\sqrt{\lambda}}}}{2} \quad (23)$$

$$\frac{\lambda \left\| q^{k+1} - q^k \right\|_{Z_1^k}^{2^{\lambda\sqrt{\lambda}}}}{2} \quad (24)$$

$$\|\varpi\|_Z = \sqrt{\langle Z_\varpi, \varpi \rangle} \quad (25)$$

where $Z$ is the positive definite matrix[3]. So the modified Eq. 18 and Eq. 19 are:

$$p^{k+1} \in arg\,min_p \left\{ \frac{\lambda \| p - p^k \|_{Z_1^k}^{2^{\lambda\sqrt{\lambda}}}}{2} + R(p) + \langle \rho^k, \mathrm{W}_k p \rangle + \frac{\daleth \| \mathrm{W}_k p - l + \mathrm{W}_k p^k - T(p^k, q^k) \|_2^2}{2} \right\} \quad (26)$$

$$q^{k+1} \in arg\,min_q \left\{ \frac{\lambda \| q - q^k \|_{Z_2^k}^{2^{\lambda\sqrt{\lambda}}}}{2} + S(q) + \langle \rho^k, \mathrm{T}_k q \rangle + \frac{\daleth \| \mathrm{T}_k q - l + \mathrm{T}_k q^k - T(p^{k+1}, q^k) \|_2^2}{2} \right\} \quad (27)$$

where $\mathrm{W}_k = \vartheta_p T(p^k, q^k)$, $\mathrm{T}_k = \vartheta_q T(p^{k+1}, q^k)$ and $\vartheta_p = {}^{\vartheta T}/_{\vartheta p} = {}^{dT}/_{dp}$

To obtain the proximity operator, we define:

$$Z_1^k = \zeta_1^k I - \daleth W_k^* \mathrm{W}_k \quad \left(\zeta_1^k < \frac{1}{\daleth \|\mathrm{W}_k\|^2}\right) \quad (28)$$

$$Z_2^k = \zeta_2^k I - \daleth T_k^* T_k \quad \left(\zeta_2^k < \frac{1}{\daleth \|\mathrm{T}_k\|^2}\right) \quad (29)$$

and then we can obtain:

[3] "A positive definite matrix is a symmetric matrix $A$ for which all eigenvalues are positive" [27].

$$\boldsymbol{p}^{k+1} = \frac{(\boldsymbol{p}^{k} - \zeta_1^k \boldsymbol{W}_k^* \times (2\boldsymbol{\rho}^k - \boldsymbol{\rho}^{k-1}))}{(\boldsymbol{I} + \zeta_1^k \boldsymbol{\vartheta R})} \quad (30)$$

$$\boldsymbol{q}^{k+1} = \frac{(\boldsymbol{q}^{k} - \zeta_2^k \boldsymbol{T}_k^* \times \left(\boldsymbol{\rho}^k + \daleth(\boldsymbol{T}(\boldsymbol{p}^{k+1}, \boldsymbol{q}^k) - \boldsymbol{l})\right))}{(\boldsymbol{I} + \zeta_2^k \boldsymbol{\vartheta S})} \quad (31)$$

based on Eq. 29 and Eq. 30, the proximity operator can be defined as:

$$\frac{\boldsymbol{\varpi}}{(\boldsymbol{I} + \boldsymbol{\alpha\vartheta R})} = \boldsymbol{arg\ min_p}\{2\boldsymbol{\alpha R}(\boldsymbol{p}) + \|\boldsymbol{p} - \boldsymbol{\varpi}\|_2^2\} \quad (32)$$

Fig.2 (c) represents the filtered depth map by using the PADMM.

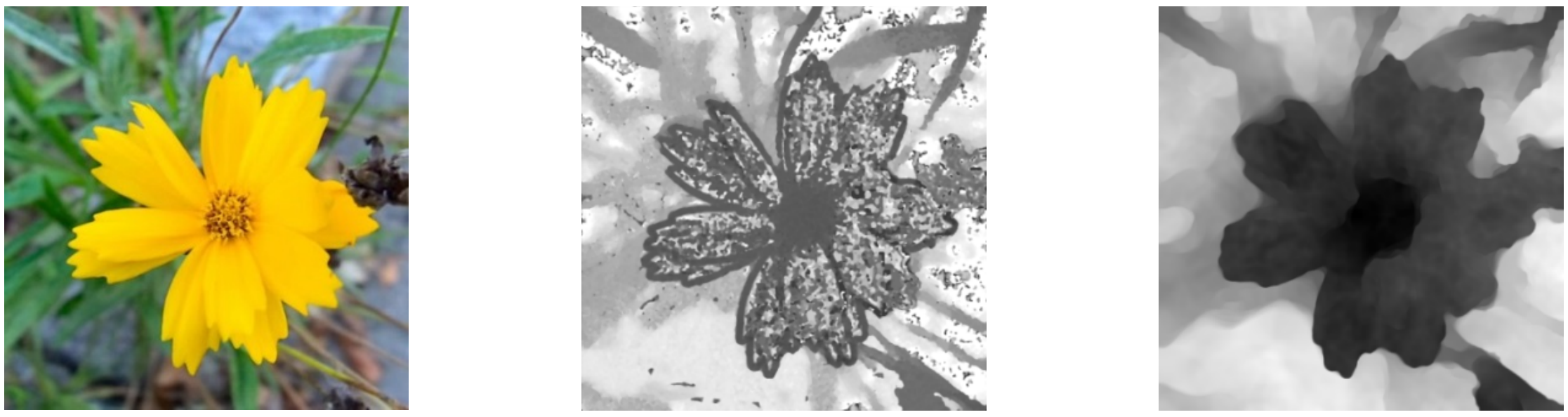

a. A frame from the focus stack | b. Initial depth map | c. Filtered depth map using PADMM

**Figure 2. The performance of the PADMM on filtering the initial depth map**

## 4 Experiments and Evaluation

For evaluation purposes, 21 sets of focal stack images by [28] are used. The focal stacks are captured using a Lytro camera which is equipped with an array of 360 × 360 micro lenses mounted on an 11 MP sensor. The images are down-sampled by the factor of 3 before applying the PADMM. Joint Bilateral Upsampling [29] is employed to up-sample the low-resolution depth map to the original size. Afterwards, the up-sampled depth map is refined using the post-processing algorithm presented in [30]. For the first part of the evaluation, the depth maps generated by the proposed method are compared against the method presented by Moeller, *et al.* [6], Helicon Focus [31] and Zerene Stacker [32]. Numerical comparison of these results is a challenging task as there is no ground truth and publicly available dataset, so the depth maps are compared visually. Fig. 3 shows the generated depth maps by the proposed method, Moeller, *et al.* [6], Helicon Focus [31] and Zerene Stacker [32]. Fig.3 (a) shows the case that the depth maps computed by Moeller, *et al.* [6], Helicon Focus [31] and Zerene Stacker [32] are missing a corner of an object and some parts of the background depth information are mixed with foreground depth values. Fig.3 (b) illustrates the scenario where the depth maps by Moeller, *et al.* [6], Helicon Focus [31] and Zerene Stacker [32] are suffering from inaccurate depth values on an object's surface. Also similar to the previous example, the background depth information is mixed with foreground depth values. Fig.3 (c) represents the case where the depth maps by Moeller, *et al.* [6], Helicon Focus [31] and Zerene Stacker [32] are not following the edges on the object's boundary. This might cause a problem in segmentation and synthetic defocus application. To find more visual results and the higher resolution version of the images presented in Fig. 3 please refer to Appendix 1 in the supplemental material.

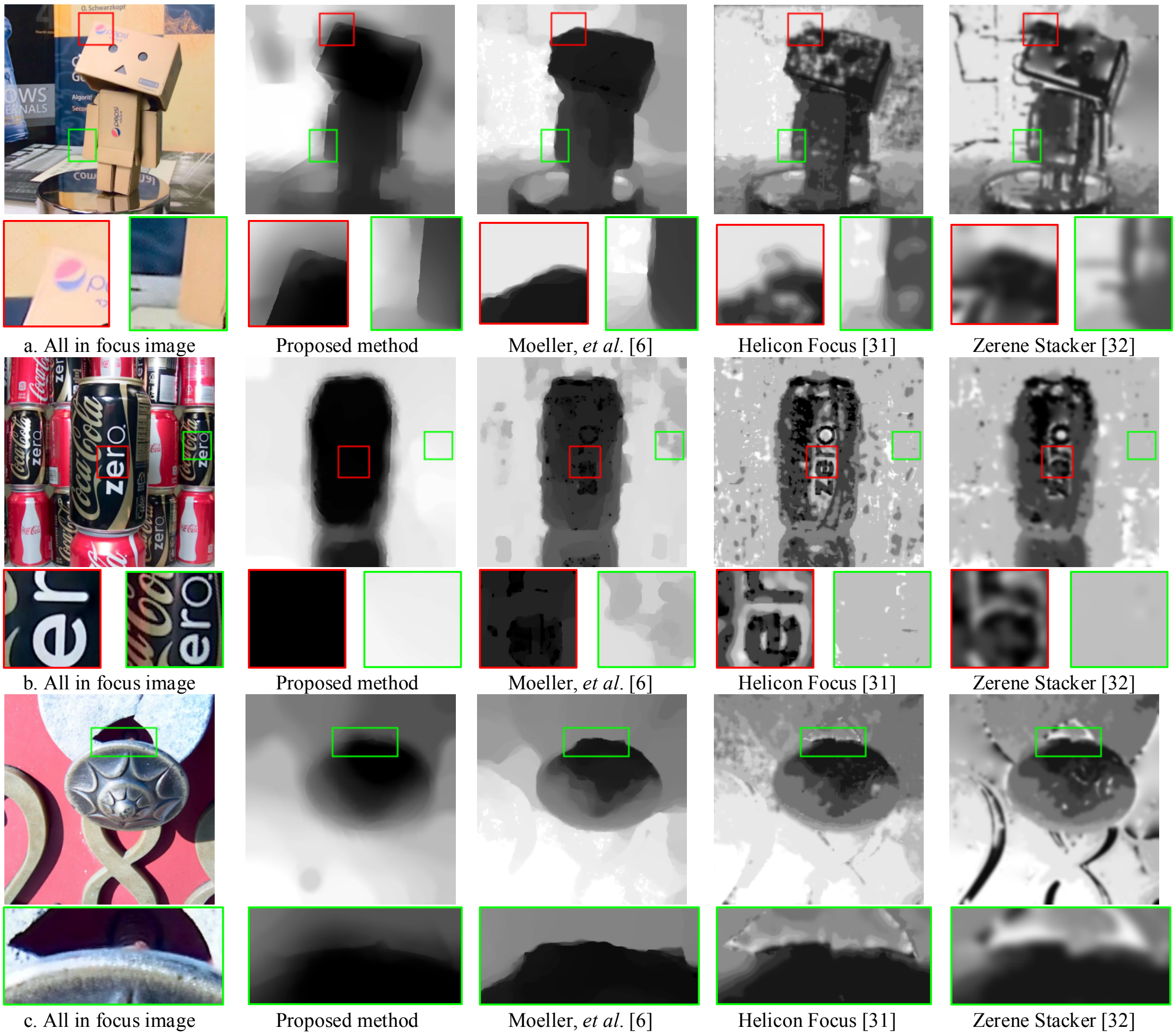

**Figure 3. The comparison of the depth maps computed by the proposed method and Moeller, *et al.* [6], Helicon Focus [31] and Zerene Stacker [32]**

To determine the performance of the generated depth maps for synthetic defocus applications, we applied hexagon shaped uniform distributed blur, based on the depth layers. Fig. 4 illustrates the synthetic defocus generated based on the depth maps presented in Fig.3 (b). Frontal object and the background are chosen as two focal points for each sample. As it is shown in Fig. 4, faulty depth values can cause artifacts in applications such as synthetic defocus and post-capture refocusing. The synthetic defocus of Fig.3 (a) and Fig.3 (c) are presented in the supplemental material.

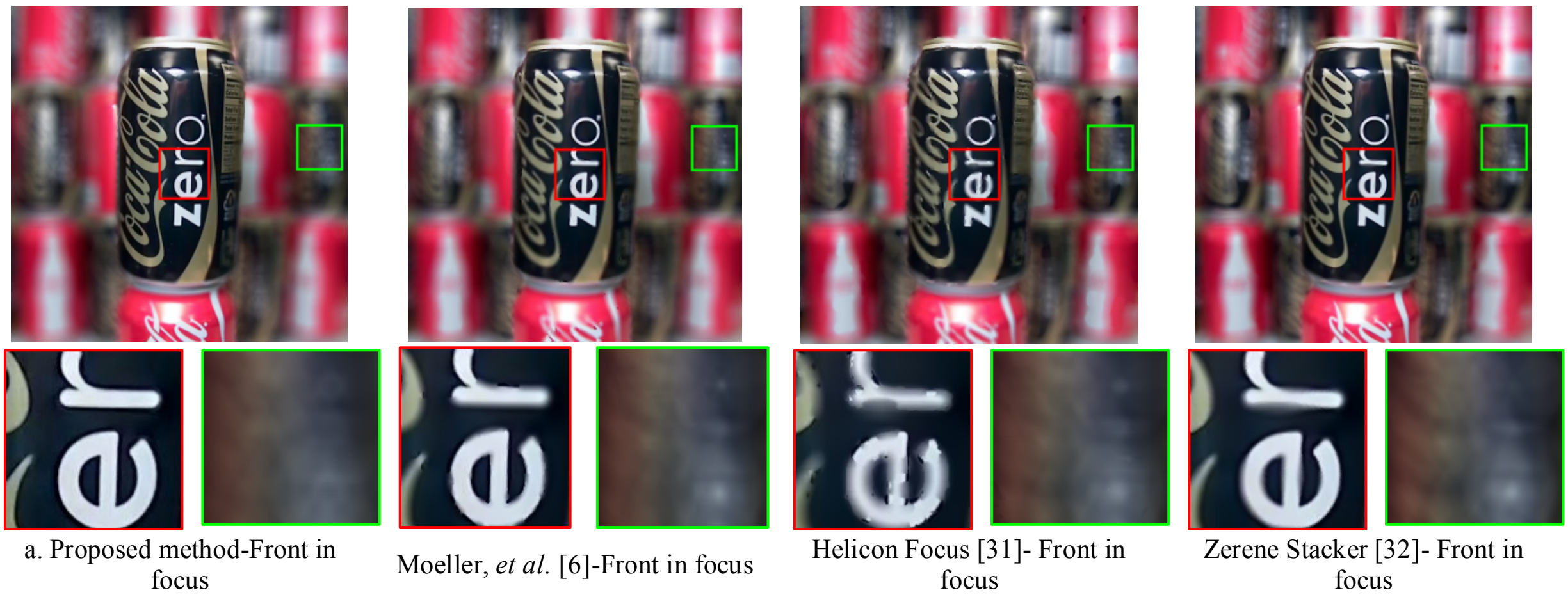

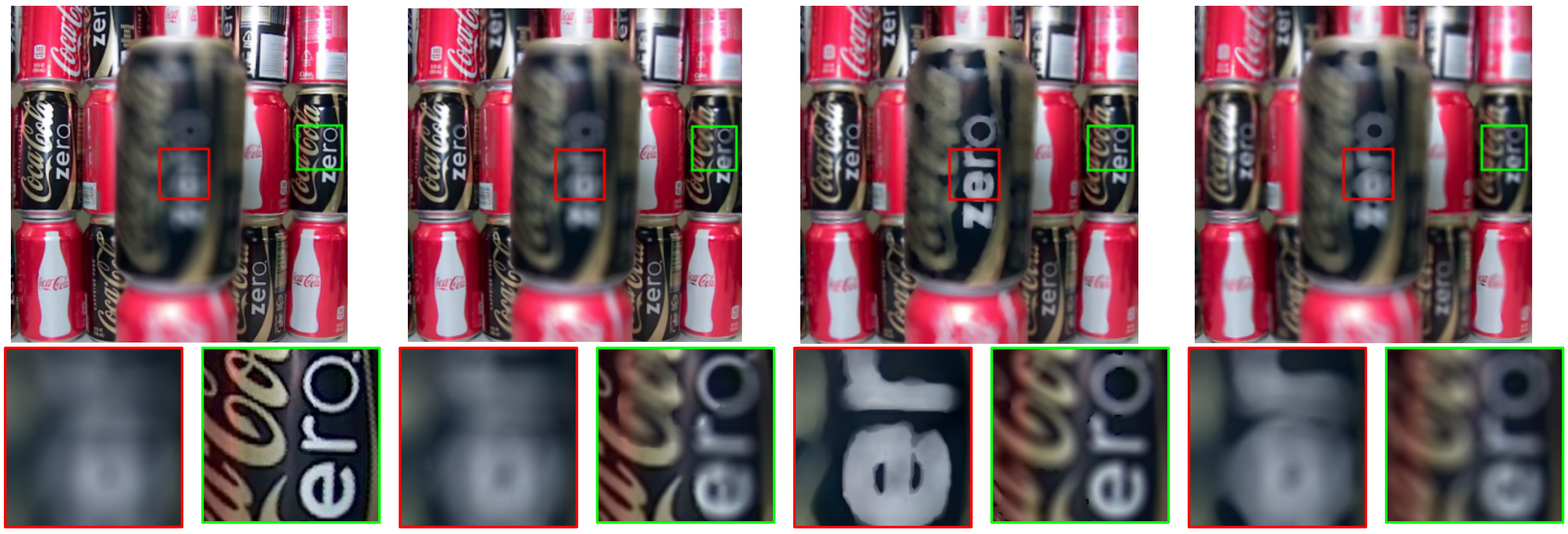

b. Proposed method-Background in focus | Moeller, *et al.* [6]-Background in focus | Helicon Focus [31]-Background in focus | Zerene Stacker [32]-Background in focus

**Figure 4. Refocusing using the recovered depth map and all in focus image**

Fig. 5 illustrates the analysis of the 3D model generated based on the depth map from the proposed method. Fig.5 (a), Fig.5 (b) and Fig.5 (c) present the all in focus image, the depth map estimated from the corresponding focal stack and the 3D color mesh generated based on the depth map, respectively. Fig.5 (d) represents the rasterized color-coded 3D model from the proposed method. The color-coded model indicates how accurate the proposed method is in terms of establishing depth levels. The transition from red to blue presents the areas which are closer and far from the camera. Fig.5 (e) shows the 3D normals of the reconstructed surface calculated using the method presented in [33]. By looking at 3D normals one can determine the smoothness of the depth values estimated by the proposed framework.

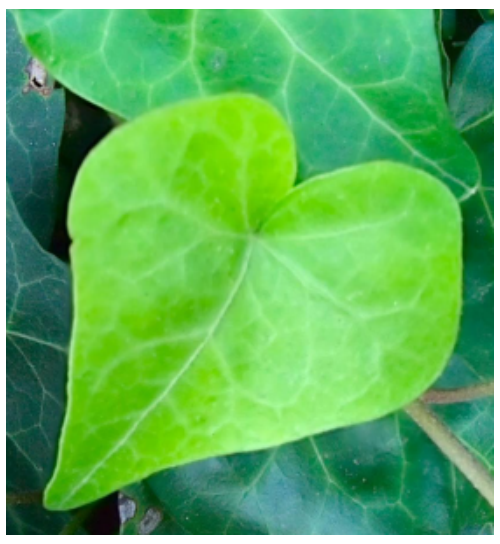

a. All in focus frame of a focal stack containing 12 images

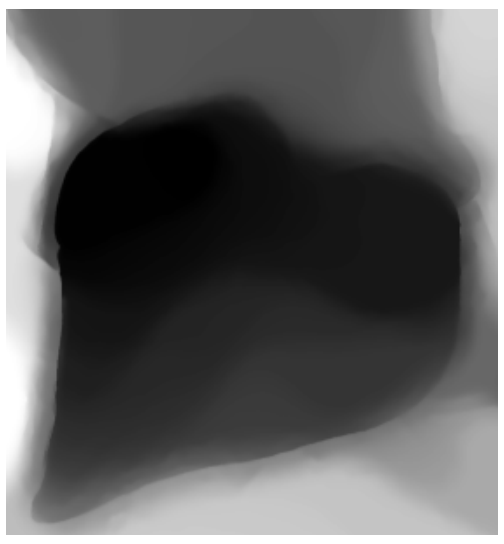

b. Estimated depth map. Dark pixels represent closer regions to the camera.

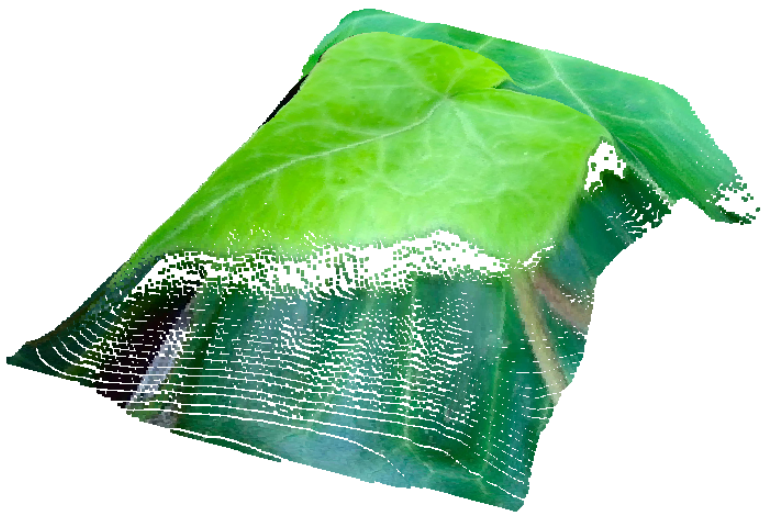

c. Proposed method – 3D color mesh

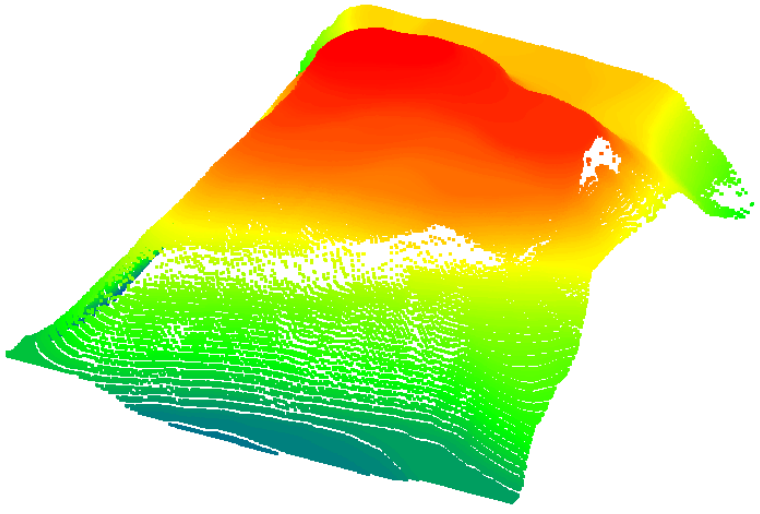

d. Proposed method – Rasterized 3D color-coded depth

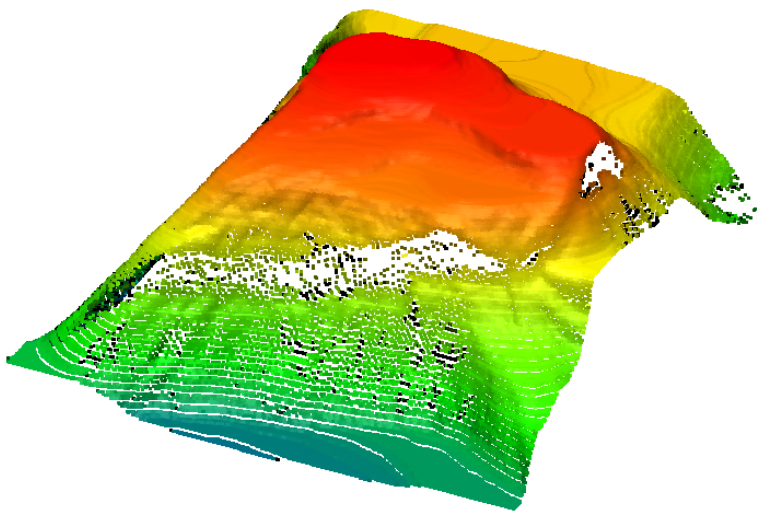

e. Proposed method – 3D Normals

**Figure 5. 3D visualizations of the depth map estimated by the proposed method. a: All in-focus image. b: Estimated depth map. c: 3D color mesh. d: Rasterized 3D color-coded depth. e: 3D normals**

At the second part of the experiment, the performance of the proposed PADMM is compared against 5 other optimization methods including Fast Iterative Shrinkage-Thresholding Algorithm (FISTA) [34], Classical Forward-Backward [35], Forward-Backward Splitting (FBS) [36], Accelerated FBS+Restart [37-39] and Adaptive Stepsize Selection FBS [39, 40]. The mean and standard deviation of the residual norm for each optimization method are illustrated in Fig. 6. The maximum number of iterations and the regularization parameter are set to 300 and 0.7 for all the methods respectively. As shown in Fig. 6, the presented PADMM optimization method, results in lower convergence error in comparison to other methods.

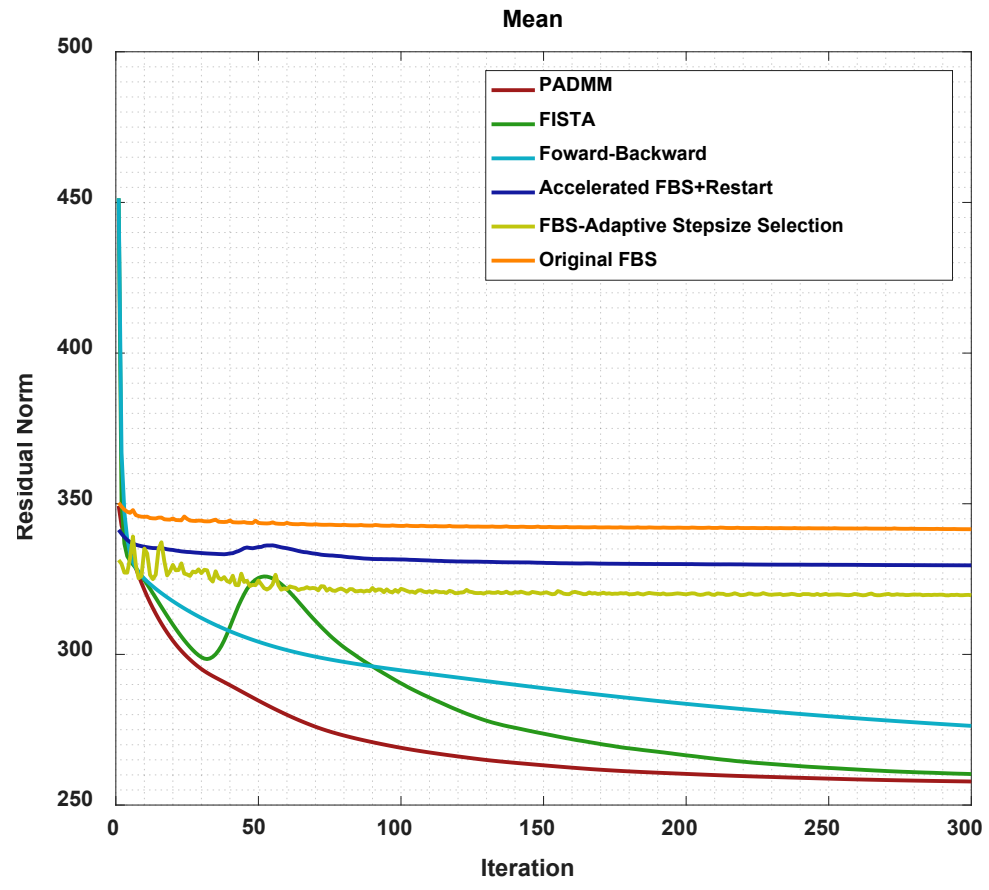

**Figure 6. Mean and Standard Deviation of the residual norm for PADMM and other 5 optimization methods**

The numerical information related to the convergence of the PADMM and Moeller, *et al.* [6] is presented as the decay of energy in a logarithmic form in Fig.7 (a) and Fig.7 (b) respectively. As it is shown in Fig.7 (a) the convergence of PADMM happens around the iteration 226 and it reaches 0.01 as the decay of energy, while the function presented by Moeller, *et al.* [6] around the same iteration reaches to the decay of 3.6 and it is still not converged. The better value of the decay of energy within the low number of iterations shows the superior performance of the proposed PADMM.

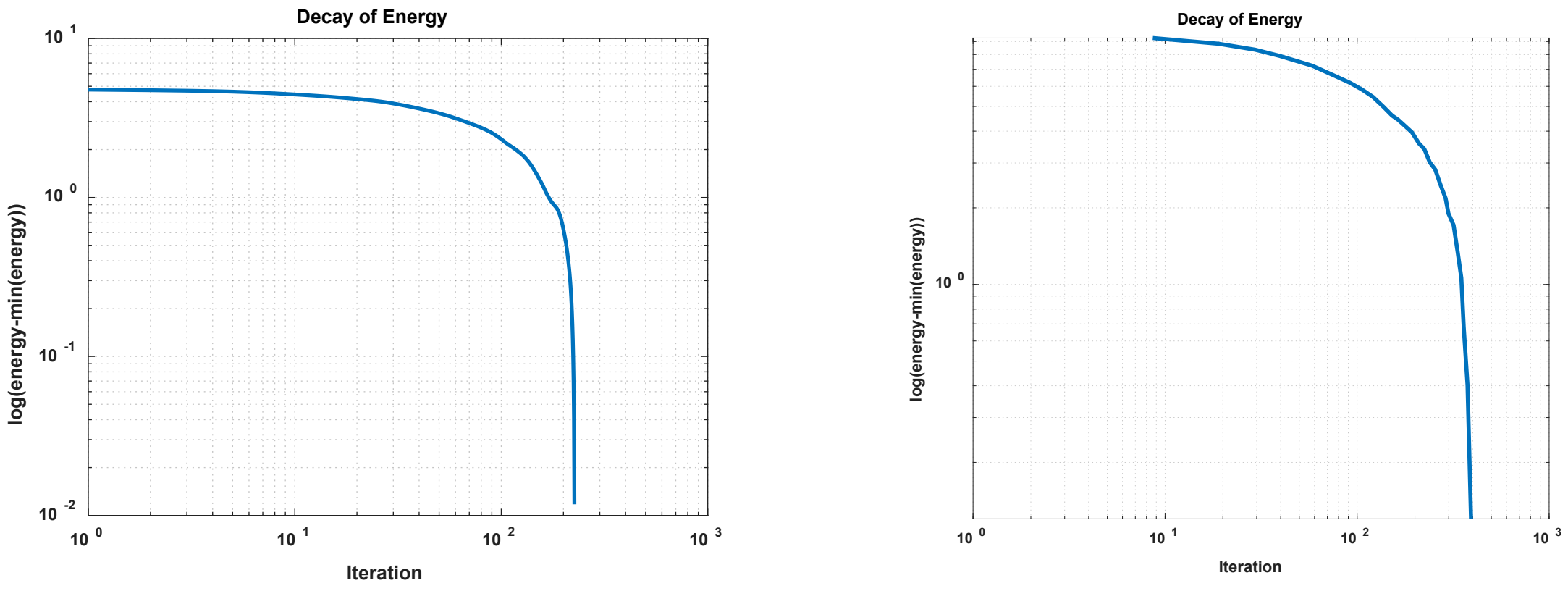

a. Convergence of PADMM as the decay of energy    b. Convergence of Moeller, *et al.* [6] as decay of energy

**Figure 7. Numerical comparison – Convergence of PADMM against Moeller, *et al.* [6]**

The third part of the comparison is done against the method proposed by Suwajanakorn, *et al.* [3]. The reason that we performed a separate comparison against this method is not having access to the code of the algorithm. The authors of [3] kindly provided the focal stacks and the depth results published in their paper. Fig. 8 illustrates the comparison of the depth maps computed by the proposed method and Suwajanakorn, *et al.* [3]. Fig.8 (a) represents the case where the depth map computed by Suwajanakorn, *et al.* [3] is suffering from inaccurate depth values on a reflective surface and some other objects' surface while the depth map by the proposed method covered these issues. The depth map by Suwajanakorn, *et al.* [3] in Fig.8 (b) shows a similar issue to the previous example, uncertain depth values along an object's edges and surface. Fig.8 (c) shows the similar issues of the reflective surfaces and inaccurate edges which have been solved by the proposed method. However, the blue highlighted part in Fig.8 (c) illustrates the case where the proposed method computed a patch of uncertain depth values on the background level. It is also worth pointing out the advantage of the method by Suwajanakorn, *et al*. [3] in computing longer depth range than the proposed method.

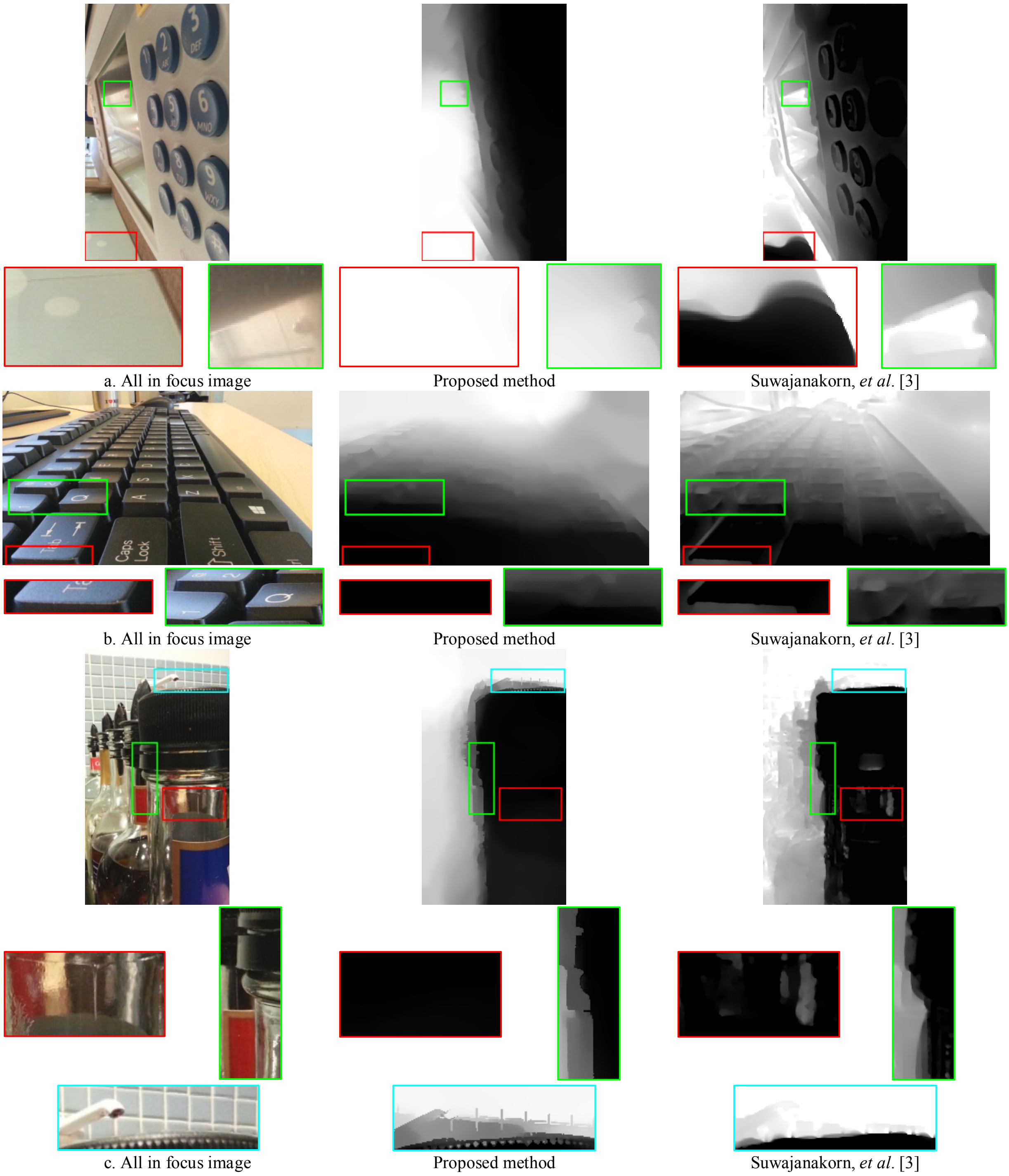

**Figure 8. The comparison of the depth maps computed by the proposed method and Suwajanakorn, *et al*. [3]**

## 5 Conclusion

In this paper, a modified version of PADMM optimization method is proposed to perform on depth from the focal stack and synthetic defocus application. The proposed method is applied on a sequence of images produces by a camera with hypothetical focus and aperture values to generate the depth map. The proposed technique satisfies the constraint of the state of the art method such as uncertain depth values on objects' surface, mixed depth values on different layers of background and foreground, missed depth information on an object's boundaries which cause faulty edges and corners in the depth map.

The method is evaluated in 2 parts. First, the generated depth maps with the correspondent defocused images are compared against a recent studied method. 21 sets of focal stack images are used in this comparison and all the parameters are set equally in both methods.

The second part of the evaluation is done to determine the performance of the proposed optimization technique in comparison to 5 other algorithms.

The results of both parts of the evaluation show that the proposed framework and modified PADMM doesn't have the best yet better performance than the recent depth from the focal stack and optimization methods.

The high structural accuracy of the depth map generated by the proposed method gives the smartphone users the ability to refocus post-capture images accurately without the need to change the aperture size. The method has been implemented in Matlab R2016a on a device equipped with Intel i7-5600U @ 2.60GHz CPU and 16 GB RAM. The computational time of the modified PADMM optimization is ~1.5 second and the whole process, from initializing the focal stack to final refined depth map takes ~53 seconds on an image with 1080×1080 pixels resolution. In our future work, we plan to implement the proposed algorithm as a smartphone application. However, despite the performance and accuracy of the studied method, there is still the computational time of this technique which has to be considered as the trade-off.